\documentclass{article}

    \usepackage[final]{neurips_2024}

\usepackage[utf8]{inputenc} 
\usepackage[T1]{fontenc}    
\usepackage{hyperref}       
\usepackage{url}            
\usepackage{booktabs}       
\usepackage{amsfonts}       
\usepackage{nicefrac}       
\usepackage{microtype}      
\usepackage{xcolor}         

\usepackage{float}
\usepackage{xspace}
\usepackage{graphicx}
\usepackage{enumitem}
\usepackage{amsmath}
\usepackage{multirow}
\usepackage{pifont}
\usepackage{xcolor}
\usepackage{colortbl}
\usepackage{wrapfig}
\usepackage{subcaption}
\def\ours{{UniSeg3D}}

\newcommand{\mysection}[1]{\noindent\textbf{#1.}}

\definecolor{isabelline}{RGB}{244, 240, 236}
\definecolor{kaiming-green}{RGB}{57,181,74} 
\hypersetup{
    colorlinks=true,
    linkcolor=red,
    filecolor=magenta,      
    urlcolor=magenta,
    citecolor=kaiming-green
}

\title{A Unified Framework for 3D Scene Understanding}

\renewcommand\footnotemark{}
\author{Wei Xu$^\ast$, Chunsheng Shi$^\ast$, Sifan Tu, Xin Zhou, \\ \textbf{Dingkang Liang, Xiang Bai$^\dagger$}\thanks{\footnotesize{$^\ast$Equal contribution. $^\dagger$Corresponding author.}}\\
	Huazhong University of Science and Technology
	\\
    \texttt{\{wxu2023, csshi, dkliang, xbai\}@hust.edu.cn}
}

\begin{document}

\maketitle

{%
\begin{figure}[H]
\hsize=\textwidth
\centering
\includegraphics[width=1.0\linewidth]{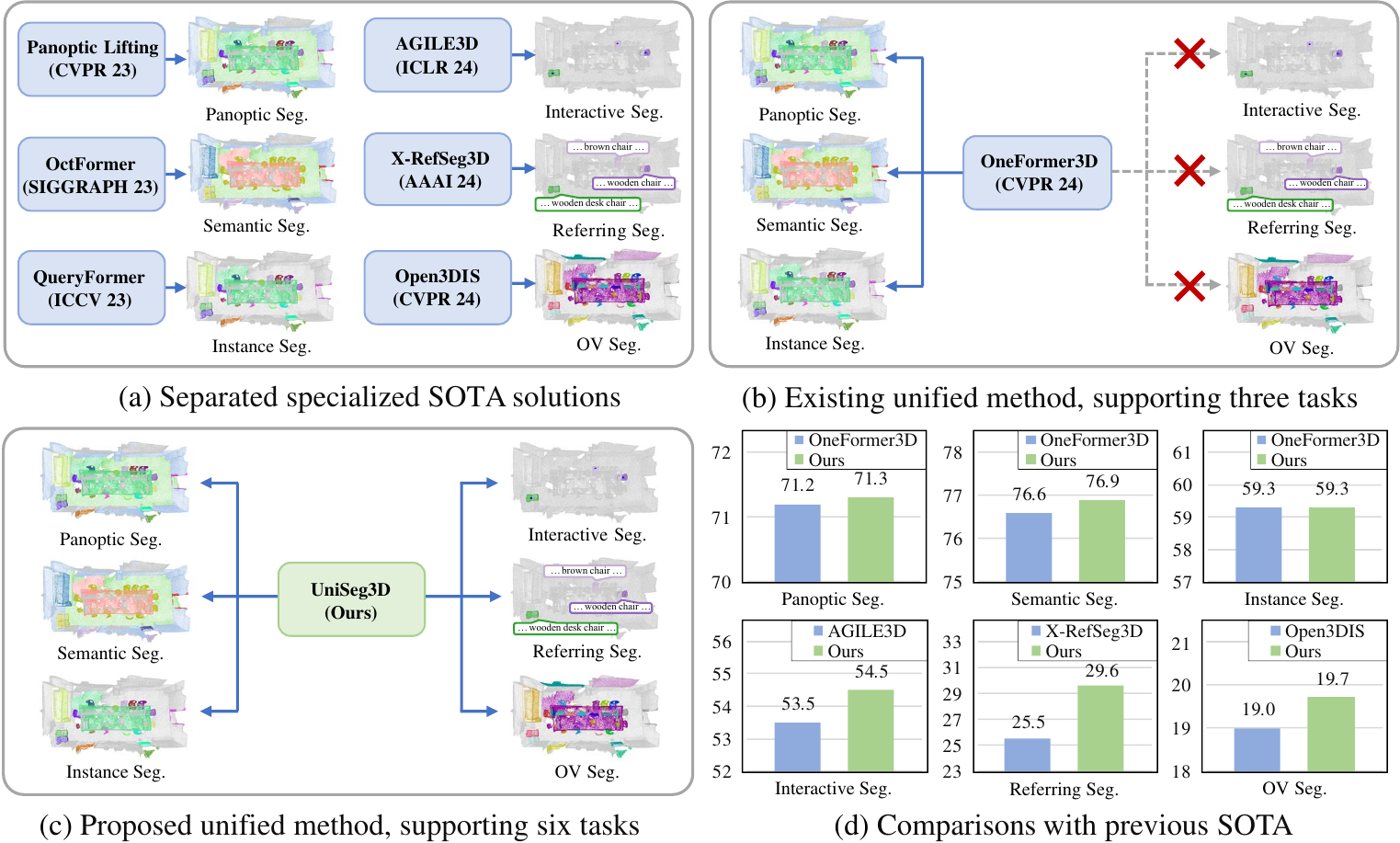}
\caption{
Comparisons between the proposed method and current SOTA approaches specialized for specific tasks. (a) Representative specialized approaches on six tasks. (b) OneFormer3D, a recent unified framework, achieves SOTA performance on three generic segmentation tasks in one inference. (c) The proposed unified framework achieves six tasks in one inference. (d) Our method outperforms current SOTA approaches across six tasks involving two modalities using a single model.
}
\label{fig:intro}
\end{figure}
}

\begin{abstract}
We propose UniSeg3D, a unified 3D scene understanding framework that achieves panoptic, semantic, instance, interactive, referring, and open-vocabulary segmentation tasks within a single model. Most previous 3D segmentation approaches are typically tailored to a specific task, limiting their understanding of 3D scenes to a task-specific perspective. In contrast, the proposed method unifies six tasks into unified representations processed by the same Transformer. It facilitates inter-task knowledge sharing, thereby promoting comprehensive 3D scene understanding. To take advantage of multi-task unification, we enhance performance by establishing explicit inter-task associations. Specifically, we design knowledge distillation and contrastive learning methods to transfer task-specific knowledge across different tasks. Experiments on three benchmarks, including ScanNet20, ScanRefer, and ScanNet200, demonstrate that the UniSeg3D consistently outperforms current SOTA methods, even those specialized for individual tasks. We hope UniSeg3D can serve as a solid unified baseline and inspire future work. Code and models are available at \url{https://dk-liang.github.io/UniSeg3D/}.
\end{abstract}

\section{Introduction}

3D scene understanding has been a foundational aspect of various real-world applications~\cite{chen2024sugar,zhang2023sam3d,jaritz2019multi}, including robotics, autonomous navigation, and mixed reality.
Among the 3D scene understanding tasks, 3D point cloud segmentation is a crucial component.
Generic 3D point cloud segmentation contains panoptic, semantic, and instance segmentation (PS/SS/IS) tasks~\cite{narita2019panopticfusion,qi2017pointnet++,wu2019pointconv,yi2019gspn,kolodiazhnyi2024top}, which segment classes annotated in the training set. As a complement, 3D open-vocabulary segmentation (OVS) task~\cite{peng2023openscene,takmaz2024openmask3d, huang2023openins3d} segments open-vocabulary classes of interest. Another group of works study to utilize user priors. In particular, 3D interactive segmentation task~\cite{kontogianni2023interactive,yue2023agile3d} segments instances specified by users. 3D referring segmentation task~\cite{huang2021text,qian2024x, 3dstmn, 3DGRES} segments instances described by textual expressions. The above mentioned tasks are core tasks in 3D scene understanding, drawing significant interest from researchers and achieving great success.

Previous studies~\cite{sun2023neuralbf,han2020occuseg,zhou2024dynamic,jiang2020pointgroup} in the 3D scene understanding area focus on separated solutions specialized for specific tasks, as shown in Fig.~\ref{fig:intro}(a). These approaches ignore intrinsic connections across different tasks, such as the objects' geometric consistency and semantic consistency.
They also fail to share knowledge biased toward other tasks, limiting their understanding of 3D scenes to a task-specific perspective. It poses significant challenges for achieving comprehensive and in-depth 3D scene understanding.
A recent exploration~\cite{kolodiazhnyi2023oneformer3d} named OneFormer3D
designs an architecture to unify the 3D generic segmentation tasks, as shown in Fig.~\ref{fig:intro}(b). 
This architecture inputs instance and semantic queries to simultaneously predict the 3D instance and semantic segmentation results. And the 3D panoptic segmentation is subsequently achieved by post-processing these predictions. It is simple yet effective.
However, this architecture fails to support the 3D interactive segmentation, 3D referring segmentation, and OVS tasks,
which provide complementary scene information, including user priors and open-set classes, 
should be equally crucial in achieving 3D scene understanding as the generic segmentation tasks.
This leads to a natural consideration that \textit{if these 3D scene understanding tasks can be unified in a single framework?}

A direct solution is to integrate the separated methods into a single architecture. However, it faces challenges balancing the customized optimizations specialized for the specific tasks involved in these methods.
Thus, we aim to design a simple and elegant framework without task-specific customized modules.
This inspires us to design the \ours, a unified framework processing six 3D segmentation tasks in parallel.
Specifically, we use queries to unify representations of the input information. 
The 3D generic segmentation tasks and the OVS task, which only input point cloud without human knowledge, thus can be processed by sharing the same workflow without worrying about prior knowledge leakage. We use a unified set of queries to represent the four-task features for simplification. The interactive segmentation inputs visual point priors to condition the segmentation.
We represent the point prompt information by simply sampling the point cloud queries, thereby avoiding repeated point feature extraction. The referring segmentation inputs textual expressions, which persist in a modality gap with the point clouds and are hard to unify in the previous workflows. To minimize the time consumption, we design a parallel text prompt encoder to extract the text queries.
All these queries are decoded using the same mask decoder and share the same output head without the design of task-specific customized structures.

We further enhance performance by taking advantage of the multi-task design.
In particular, we empirically find that the interactive segmentation outperforms the rest of the tasks in mask predictions, which is attributable to reliable vision priors.
Hence, we design knowledge distillation to distill knowledge from the interactive segmentation to the other tasks. Then, we build contrastive learning between interactive segmentation and referring segmentation to connect these two tasks.
The proposed knowledge distillation and contrastive learning
promote knowledge sharing across six tasks,
%
effectively establishing associations between different tasks.
There are three significant strengths of the \ours: (1) it unifies six 3D scene understanding tasks in a single framework, as shown in Fig.~\ref{fig:intro}(c); (2) it is flexible for that can be easily extended to more tasks by simply inputting the additional task-specific queries; (3) the designed knowledge distillation and contrastive learning are only used in the training phase, optimizing the performance with no extra inference cost.

We compare the proposed method with task-specific specialized SOTA approaches~\cite{siddiqui2023panoptic,wang2023octformer,lu2023query,yue2023agile3d,qian2024x,nguyen2023open3dis} across six tasks to evaluate its performance. As shown in Fig.~\ref{fig:intro}(d), the \ours~demonstrates superior performance on all the tasks. It is worth noting that our performance on different tasks is achieved by a single model, which is more efficient than running separate task-specific approaches individually. Furthermore, the structure of \ours~is simple and elegant, containing no task-customized modules, while consistently outperforming specialized SOTA solutions,
demonstrating a desirable potential to be a solid unified baseline.

In general, our contributions can be summarized as follows: \textbf{First}, we propose a unified framework named \ours, offering a flexible and efficient solution for 3D scene understanding. It achieves six 3D segmentation tasks in one inference by a single model. To the best of our knowledge, this is the first work to unify six 3D segmentation tasks. \textbf{Second}, specialized approaches limit their 3D scene understanding to task-specific perspectives. 
We facilitate inter-task knowledge sharing to promote comprehensive 3D scene understanding. Specifically, we take advantage of multi-task unification, designing the knowledge distillation and contrastive learning methods to establish explicit inter-task associations.

\section{Related Work}

\textbf{3D segmentation.}
The generic segmentation consists of panoptic, semantic, and instance segmentation.
The panoptic segmentation~\cite{narita2019panopticfusion,wu2021scenegraphfusion} is the union of instance segmentation~\cite{he2021dyco3d,liang2021sstnet,chen2021hais,wu2022dknet,vu2022softgroup} and semantic segmentation~\cite{qian2022pointnext,qi2017pointnet++,choy2019minkowski,zhao2021pointtransformer}. It contains the instance masks from the instance segmentation and the stuff masks from the semantic segmentation. These 3D segmentation tasks rely on annotations, segmenting classes labeled in the training set. The open-vocabulary segmentation~\cite{nguyen2023open3dis,takmaz2024openmask3d} extends the 3D segmentation to the novel class. Another group of works explores 3D segmentation conditioned by human knowledge. Specifically, the interactive segmentation~\cite{kontogianni2023interactive,yue2023agile3d} segments instances specified by the point clicks. The referring segmentation~\cite{huang2021text,qian2024x} segments objects described by textual expressions.
Most previous researches~\cite{xu2020squeezesegv3,cheng2021net,lai2022stratified,liang2024pointmamba} focus on specific 3D segmentation tasks, limiting their efficiency in multi-task scenarios, such as the domotics, that require multiple task-specific 3D segmentation approaches to be applied simultaneously. This work proposes a framework to achieve the six abovementioned tasks in one inference.

\textbf{Unified vision models.}
Unified research supports multiple tasks in a single model, facilitating efficiency and attracting a lot of attention in the 2D area~\cite{qi2023unigs,li2024univs,li2024omg,jain2023oneformer}. However, rare works study the unified 3D segmentation architecture. It might be attributed to the higher dimension of the 3D data, which leads to big solution space, making it challenging for sufficient unification across multiple 3D tasks.
Recent works~\cite{hong2024unified,liu2024multi} explore outdoor unified 3D segmentation architectures, and some others~\cite{zhu20233d, huang2023ponder, irshad2024nerfmae} delve into unified 3D representations. 
So far, only one method, OneFormer3D~\cite{kolodiazhnyi2023oneformer3d}, focuses on indoor unified 3D segmentation. It extends the motivation proposed in OneFormer~\cite{jain2023oneformer} to the 3D area and proposes an architecture to achieve three 3D generic segmentation tasks in a single model.
We note that the supported tasks in OneFormer3D can be achieved in one inference through post-processing predictions of a panoptic segmentation model.
In contrast, we propose a simple framework that unifies six tasks, including not only generic segmentation but also interactive segmentation, referring segmentation, and OVS, into a single model. Additionally, we establish explicit associations between these unified tasks to promote knowledge sharing, contributing to effective multi-task unification.

\section{Methodology}

The framework of \ours~is depicted in Fig.~\ref{fig:pipeline}. It mainly consists of three modules: a point cloud backbone, prompt encoders, and a mask decoder.
We illustrate their structures in the following.

\subsection{Point Cloud Backbone and Prompt Encoders}
\label{framework:encoder}

\textbf{Point cloud backbone.} We represent the set of $N$ input points as $\mathbf{P}\in  \mathbb{R} ^{N\times 6}$, where each point is characterized by three-dimensional coordinates $x$, $y$, $z$ and three-channel colors $r$, $g$, $b$. These input points are then fed into a sparse 3D U-Net, serving as the point cloud backbone, to obtain point-wise features $\mathbf{F}\in  \mathbb{R} ^{N\times d_{in}}$, where $d_{in}$ denotes the feature dimension. Processing dense points individually in 3D scene understanding can be time-consuming. Therefore, we downsample the 3D scenario into $M$ superpoints and pool the point features within each superpoint to form superpoint features $\mathbf{F}_{s}=\left \{ \mathbf{f}_i  \right \} _{i=1}^{M}$, where each $\mathbf{f}_i \in \mathbb{R} ^{d_{in}}$ and $\mathbf{F}_{s} \in \mathbb{R} ^{M\times d_{in}}$. This procedure exhibits awareness of the edge textures~\cite{landrieu2018large} while reducing cost consumption.

\begin{figure}[t] 
\centering
\includegraphics[width=1.0\textwidth]{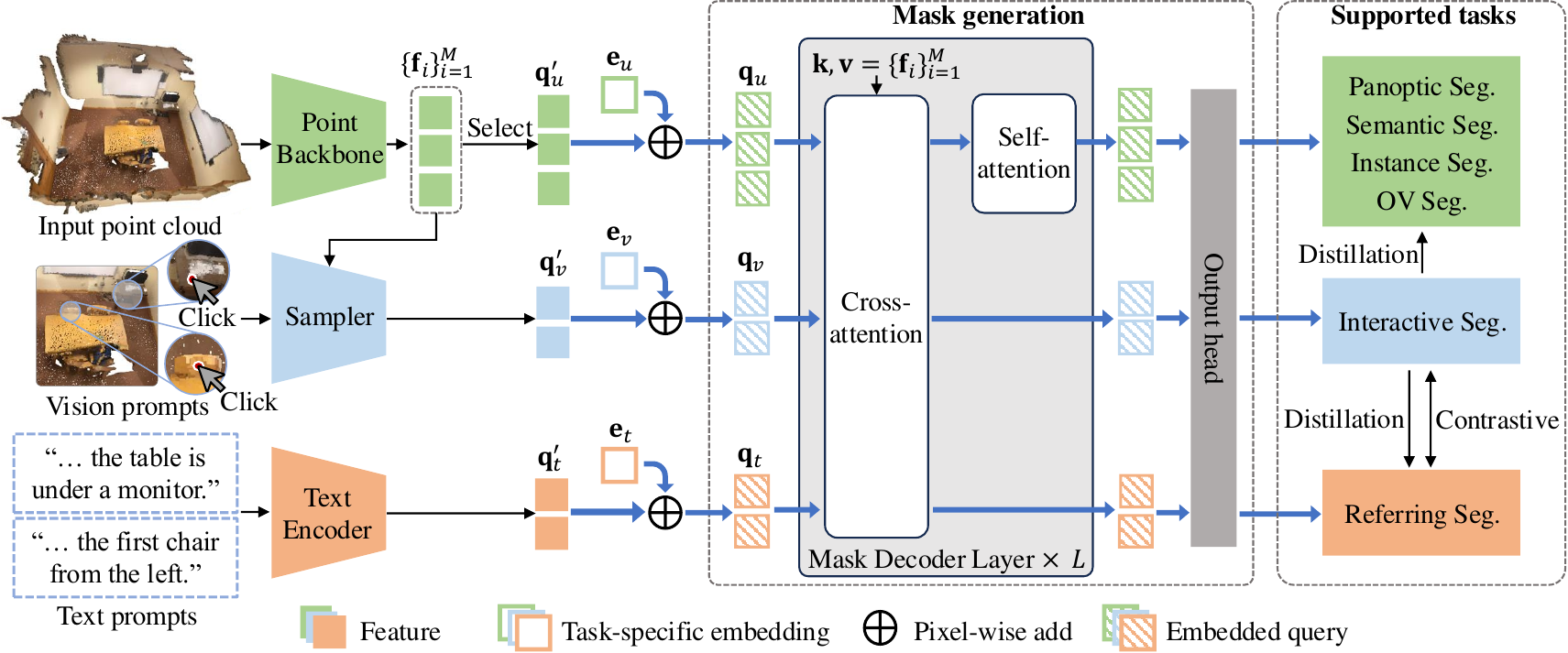}
\caption{The framework of \ours. This is a simple framework handling
six tasks
in parallel without any modules specialized for specific tasks. We take advantage of multi-task unification and enhance the performance through building associations between the supported tasks. Specifically, knowledge distillation transfers insights from interactive segmentation to the other tasks, while contrastive learning establishes connections between interactive segmentation and referring segmentation.
}
\label{fig:pipeline}
\end{figure}

\textbf{Vision prompt encoder.} Click is a kind of clear and convenient visual interaction condition that has been widely employed in previous works~\cite{kirillov2023segment,kontogianni2023interactive,yue2023agile3d}. We formulate the clicks as vision prompts, as illustrated in Fig.~\ref{fig:pipeline}. 
In practice, a click is first indicated by the spatially nearest point. Then, we sample a superpoint containing this point and employ its superpoint feature as vision prompt feature
$\mathbf{f}_{v} \in \mathbb{R}^{d_{in}}$ to represent the point prompt information, thus avoiding redundant feature extraction and maintaining feature consistency with the point clouds.

\textbf{Text prompt encoder.} \ours~is able to segment instances described by textual expressions. To process a text prompt, the initial step involves tokenizing the text sentence to obtain its string tokens $\mathbf{T} \in \mathbb{R}^{l \times c}$, where $l$ is the sentence length and $c$ represents the token dimension. These tokens are then fed into a frozen CLIP~\cite{radford2021learning} text encoder to produce a $C$-dimensional text feature $\mathbf{f}_{t} \in \mathbb{R}^{C}$. This feature is subsequently projected into the $d_{in}$ dimension using two linear layers, obtaining $\mathbf{f}_{t} \in \mathbb{R}^{d_{in}}$, aligning the dimension of the point features for subsequent processing.

\subsection{Mask Generation}
\label{framework:decoder}

We employ a single mask decoder to output predictions of six 3D scene understanding tasks. The generic segmentation and the OVS share the same input data, \textit{i.e.}, the point cloud without user knowledge. Therefore, we randomly select $m$ features from $M$ superpoint features to serve as unified queries $\mathbf{q}'_{u}\in \mathbb{R} ^{m \times d_{in}}$ for both the generic segmentation and OVS tasks. During training, we set $m<M$ to reduce computational costs, while for inference, we set $m=M$ to enable the segmentation of every region.

The prompt information is encoded into prompt features as discussed in Sec.~\ref{framework:encoder}. We employ the prompt features as the prompt queries, which can be written as: $\mathbf{q}'_{v} = \left \{ \mathbf{f}_{v,i} \right \}_{i = 1}^{K_{v}}$, $\mathbf{q}'_{t} = \left \{ \mathbf{f}_{t,i} \right \}_{i = 1}^{K_{t}}$, where $\mathbf{q}'_{v}\in \mathbb{R} ^{K_{v} \times d_{in}}$, $\mathbf{q}'_{t}\in \mathbb{R} ^{K_{t} \times d_{in}}$.
$K_{v}$ and $K_{t}$ are the number of the point and text prompts, respectively. $\mathbf{q}'_{u}$, $\mathbf{q}'_{v}$, $\mathbf{q}'_{t}$ are three types of queries containing information from various aspects. Feeding them forward indiscriminately would confuse the mask decoder for digging task-specific information. Thus, we add task-specific embeddings $\mathbf{e}_u$, $\mathbf{e}_v$, and $\mathbf{e}_t$ before further processing:
\begin{align}
\mathbf{q}_{u}=\mathbf{q}'_{u}+\mathbf{e}_u,\quad\mathbf{q}_{v}=\mathbf{q}'_{v}+\mathbf{e}_v,\quad\mathbf{q}_{t}=\mathbf{q}'_{t}+\mathbf{e}_t,
\end{align}
where $\mathbf{e}_u\in  \mathbb{R} ^{d_{in}}$, $\mathbf{e}_v\in  \mathbb{R} ^{d_{in}}$, $\mathbf{e}_t\in  \mathbb{R} ^{d_{in}}$, and are broadcasted into $\mathbb{R} ^{m \times d_{in}}$, $\mathbb{R} ^{K_{v} \times d_{in}}$, and $\mathbb{R} ^{K_{t} \times d_{in}}$, respectively.
The mask decoder comprises $L$ mask decoder layers, which contain self-attention layers integrating information among queries. Prompt priors are unavailable for generic segmentation during inference. Therefore, in the training phase, we should prevent the human knowledge from leaking to the generic segmentation. In practice, the prompt queries are exclusively fed into the cross-attention layers.
Output queries of the last mask decoder layer are sent into an output head, which consists of MLP layers to project dimensions $d_{in}$ of the output queries into $d_{out}$.
In general, the mask generation process can be formally defined as:
\begin{align}
\mathbf{F}_{out}=\mathrm {MLP}\left(\mathrm { MaskDecoder }\left(\mathbf{q}=\mathrm{Concat} \left(\mathbf{q}_{u},\mathbf{q}_{v},\mathbf{q}_{t}\right);\mathbf{k}=\mathbf{F}_{s};\mathbf{v}=\mathbf{F}_{s}\right)\right),
\end{align}
where $\mathbf{F}_{out}=\left\{\mathbf{f}_{out,i}\right\}_{i=1}^{m+K_{v}+K_{t}}$ represents output features, with $\mathbf{f}_{out,i}\in\mathbb{R}^{d_{out}}$ and $\mathbf{F}_{out}\in\mathbb{R}^{(m+K_{v}+K_{t})\times d_{out}}$.

Subsequently, we can process the output features to obtain the class and mask predictions. For class predictions, a common practice involves replacing class names with class IDs~\cite{kolodiazhnyi2023oneformer3d}. However, for our method to support referring segmentation, the class names are crucial information that should not be overlooked. 
Hence, we encode the class names into text features $\mathbf{e}_{cls} \in \mathbb{R}^{K_{c} \times d_{out}}$ using a frozen CLIP text encoder and propose to regress the class name features instead, where $K_{c}$ denotes the number of categories. 
Specifically, we formulate the mask predictions $\mathbf{mask}_{pred}$ and class predictions $\mathbf{cls}_{pred}$ as follows:
\begin{align}
\mathbf{mask}_{pred}=\mathbf{F}_{out}\times\mathrm{MLP}\left (\mathbf{F}_{s}\right )^{\top},\quad\mathbf{cls}_{pred}=\mathrm{Softmax} \left ( \mathbf{F}_{out}\times\mathbf{e}_{cls}^{\top} \right ),
\end{align}
where $\mathbf{mask}_{pred}=\left\{\mathbf{mask}_{i}\right\}_{i=1}^{m+K_{v}+K_{t}}$
and $\mathbf{cls}_{pred}=\left\{\mathbf{cls}_{i}\right\}_{i=1}^{m+K_{v}+K_{t}}$, with $\mathbf{mask}_{pred}\in \mathbb{R}^{(m+K_{v}+K_{t})\times m}$ and $\mathbf{cls}_{pred}\in \mathbb{R}^{(m+K_{v}+K_{t})\times K_{c}}$.
$\mathbf{mask}_{i} \in \mathbb{R}^{m}$ and $\mathbf{cls}_{i} \in \mathbb{R}^{K_{c}}$ represent the mask outcome and category probability predicted by the $i$-th query, respectively. The $\mathrm{MLP}$ projects $\mathbb{R}^{d_{in}}$ into $\mathbb{R}^{d_{out}}$.
Given that $\mathbf{mask}_{pred}$ and $\mathbf{cls}_{pred}$ are derived from superpoints, we map the segmentation outputs for each superpoint back to the input point cloud to generate point-wise mask and class predictions.

\begin{figure}[t] 
\centering
\includegraphics[width=1.0\textwidth]{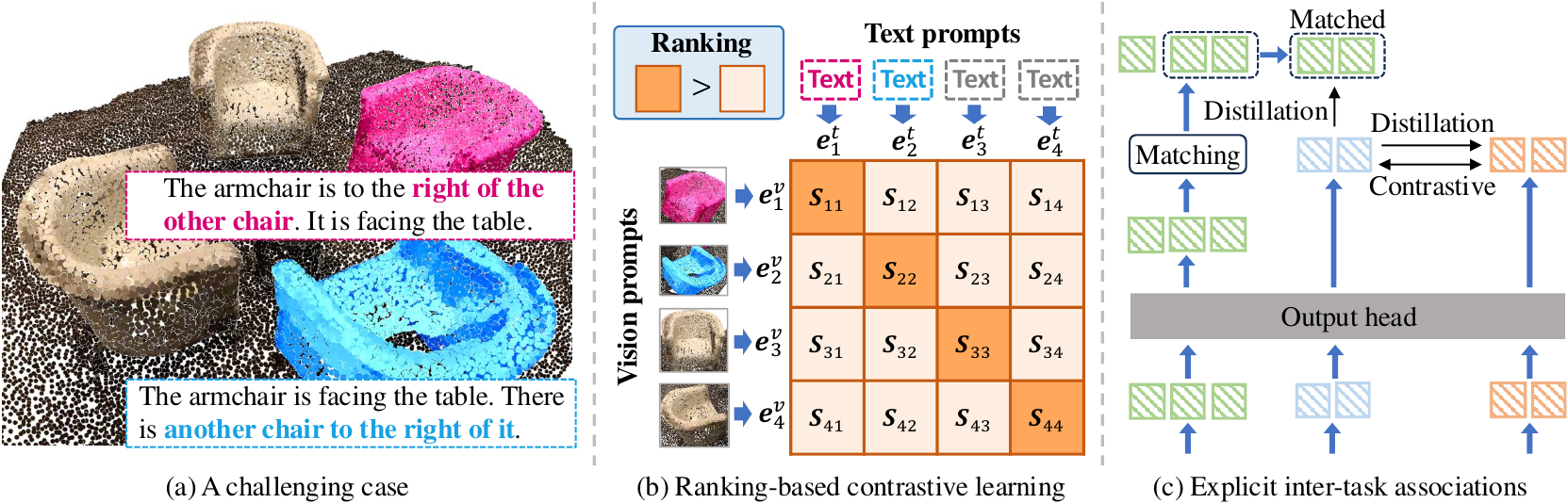}
\caption{Illustration of the inter-task association. (a) A challenging case requiring the distinction of textual positional information within the expressions. (b) A contrastive learning matrix for vision-text pairs, where a ranking rule is employed to suppress incorrect pairings. (c) Knowledge distillation across multi-task predictions.
}
\label{fig:guidance}
\end{figure}

\subsection{Explicit Inter-task Association}
\label{framework:connection}

Previous studies have overlooked the associations among 3D scene understanding tasks, resulting in task-specific approaches that fail to leverage cross-task knowledge. This limitation restricts the understanding of 3D scenes to a task-specific perspective, hindering comprehensive 3D scene understanding. We establish explicit inter-task associations to overcome these constraints.

Specifically, on the one hand, as shown in Fig.~\ref{fig:guidance}(a), the referring segmentation is challenging when multiple individuals of identical shapes are arranged adjacently. It requires the method to distinguish the location variations inserted in the text prompts, such as ``right of the other chair'' vs. ``another chair to the right of it.'' However, the modality gap between 3D points and 
linguistic texts sets significant obstructions. We propose ranking-based contrastive learning between the vision and text features to reduce the modality gap and optimize the referring segmentation.

\begin{wraptable}{r}{0.30\textwidth}
\small
\centering
\caption{Mask prediction performance of instance and interactive segmentation.}
\resizebox{0.30\textwidth}{!}{
\begin{tabular}{ cccc }
\toprule
Tasks & mIoU\\
\midrule
Instance Seg. & 68.1 \\
Interactive Seg. & \textbf{76.0} (+7.9) \\
\bottomrule
\end{tabular}
}
\label{tab:interactive}
\end{wraptable}
On the other hand, as shown in Tab.~\ref{tab:interactive}, we evaluate our baseline framework built in Sec.~\ref{framework:encoder} and Sec.~\ref{framework:decoder} on instance and interactive segmentation tasks. Essentially, the main difference between the instance and interactive segmentation is w/o or w/ vision prompts. The mIoU metric, which directly measures the quality of mask predictions, indicates that the interactive segmentation surpasses the instance segmentation by a notable margin of $7.9\%$. It suggests that the vision prompts provide reliable position priors, boosting the interactive segmentation to perform superior mask prediction performance. We design a knowledge distillation to share insights from the interactive segmentation, leveraging its superior mask prediction capability. The key to knowledge distillation is to utilize task-predicting segmentation masks of the best quality to guide the other tasks, \textit{i.e.}, using a teacher to guide students.

\subsubsection{Ranking-based Contrastive Learning}

We set the vision and text prompts specifying the same individual instances into pairs and align their pair-wise features by employing contrastive learning.

Assuming $B$ vision-text pairs within a training mini-batch, the corresponding output features are $\left \{ \mathbf{f}_{out,i}^{v} \right \}_{i=1}^{B}$ and $\left \{\mathbf{f}_{out,i}^{t} \right \}_{i=1}^{B}$, where $\mathbf{f}_{out,i}^{v}$ and $\mathbf{f}_{out,i}^{t}$ are selected from output features $\left\{\mathbf{f}_{out,i}\right\}_{i=m+1}^{m+K_{v}}$ and $\left\{\mathbf{f}_{out,i}\right\}_{i=m+K_{v}+1}^{m+K_{v}+K_{t}}$, respectively. We normalize the pair-wise vision-text output features $\left \{ \mathbf{f}_{out,i}^{v} \right \}_{i=1}^{B}$ and $\left \{\mathbf{f}_{out,i}^{t} \right \}_{i=1}^{B}$ and obtain the metric embeddings $\left \{ \mathbf{e} _{i}^{v} \right \}_{i=1}^{B} $ and $\left \{ \mathbf{e} _{i}^{t} \right \} _{i=1}^{B} $, respectively. We formulate the contrastive learning loss as $\mathcal{L}_{con}=\mathcal{L}_{v}+\mathcal{L}_{t}$, with:
\begin{align}
\mathcal{L}_{v}=-\frac{1}{B} \sum_{i=1}^{B} \log \frac{\exp \left(\mathbf{e} _{i}^{v} \cdot \mathbf{e} _{i}^{t} / \tau \right)}{\sum_{j=1}^{B} \exp \left(\mathbf{e} _{i}^{v} \cdot \mathbf{e} _{j}^{t} / \tau \right)}, ~~~\mathcal{L}_{t}=-\frac{1}{B} \sum_{i=1}^{B} \log \frac{\exp \left(\mathbf{e} _{i}^{t} \cdot \mathbf{e} _{i}^{v} / \tau \right)}{\sum_{j=1}^{B} \exp \left(\mathbf{e} _{i}^{t} \cdot \mathbf{e} _{j}^{v} / \tau \right)},
\end{align}
where $\tau $ is a learnable temperature scaling factor. The pair-wise similarity is illustrated in Fig.~\ref{fig:guidance}(b), where we denote $\mathbf{e} _{i}^{v} \cdot \mathbf{e} _{j}^{t}$ as $\mathbf{s}_{i,j}$ for simplification. To distinguish the target instances from adjacent ones with identical shapes, we introduce a ranking rule inspired by the CrowdCLIP~\cite{liang2023crowdclip} that the diagonal elements are greater than the off-diagonal elements, which can be described as:
\begin{align}
\mathcal{L}_{rank}=\frac{1}{B}\sum_{i=1}^{B}\sum_{j=1}^{B}\mathrm{max\left ( 0,\mathbf{s}_{i,j}-\mathbf{s}_{i,i} \right ) }.
\end{align}

\subsubsection{Knowledge Distillation}
As shown in Fig.~\ref{fig:guidance}(c), we transfer knowledge from the interactive segmentation task to the generic and referring segmentation tasks to guide their training phases.

\textbf{Interactive segmentation to generic segmentation task.} Define the predictions from the unified queries as $Pred_{u}=\left\{\mathbf{mask}_{i},\mathbf{cls}_{i}\right\}_{i=1}^{m}$.
We employ the Hungarian algorithm, utilizing the Dice and cross-entropy metrics as the matching cost criteria, to assign $Pred_{u}$ with the interactive segmentation labels $GT_{v}=\left\{\mathbf{mask}_{gt,i},\mathbf{cls}_{gt,i}\right\}_{i=1}^{K_{v}}$. The matched predictions are selected as positive samples $Pos_{u}=\left\{\mathbf{mask}_{pos,i},\mathbf{cls}_{pos,i}\right\}_{i=1}^{K_{v}}$. The predicted masks from the interactive segmentation can be formulated as $\mathbf{mask}_{v}=\left\{\mathbf{mask}_{i}\right\}_{i=m+1}^{m+K_{v}}$, 
where $\mathbf{mask}_{v} \in \mathbb{R}^{K_{v}\times m}$. We select the pixels with top $k\%$ scores of $\mathbf{mask}_{v}$ as learning region $\mathbf{R}$, and depict the knowledge transfer process from the interactive segmentation to the generic segmentation task as:
\begin{align}
\mathcal{L}_{v\to g}=\mathcal{L}_{BCE} \left ( \mathbf{mask}_{pos}\left( \mathbf{R} \right), \mathbf{mask}_{v}\left( \mathbf{R} \right) \right ) ,
\end{align}
where $\mathbf{mask}_{pos}\left( \mathbf{R} \right)$ and $\mathbf{mask}_{v}\left( \mathbf{R} \right)$ represent the predicted mask values within the region $\mathbf{R}$, gathering from the positive samples and the interactive segmentation predictions, respectively.

\textbf{Interactive segmentation to referring segmentation task.} Define the pair-wise class probabilities predicted by the vision and text prompt queries as $\mathbf{cls}_{v}\in \mathbb{R}^{B \times K_{c}}$ and $\mathbf{cls}_{t}\in \mathbb{R}^{B \times K_{c}}$ selected from $\left\{\mathbf{cls}_{i}\right\}_{i=m+1}^{m+K_{v}}$ and $\left\{\mathbf{cls}_{i}\right\}_{i=m+K_{v}+1}^{m+K_{v}+K_{t}}$, respectively.
We formulate a knowledge transfer process from the interactive segmentation to the referring segmentation task as:
\begin{align}
\mathcal{L}_{v\to r}=\mathcal{L}_{BCE} \left ( \mathrm{Sigmoid}\left ( \mathbf{cls}_{t} \right )  , \mathrm{Sigmoid}\left ( \mathbf{cls}_{v} \right ) \right ).
\end{align}

\subsection{Training Objectives}
\label{framework:loss}

\textbf{Open-set pseudo mask labels.}
For open-vocabulary tasks, we train models on close-set data. To enhance segmentation performance on open-set data, we use SAM3D~\cite{yang2023sam3d} to generate segmentation masks with undetermined categories as pseudo mask labels (open-set masks). While training, we assign predictions of the unified queries with ground-truth masks (close-set masks). The assigned and miss-assigned predictions are divided into positive and negative samples, respectively. The positive samples are supervised to regress the close-set masks. We match the negative samples with the pseudo mask labels and supervise the matched ones to regress the open-set masks.
Note that the SAM3D is an unsupervised method and does not rely on ground-truth annotations, eliminating worries of label leakage. 
This process is exclusively applied in the training phase, incurring no extra inference cost.

\textbf{Loss function.}
The training losses contain two components: (1) the basic losses, formulated as $\mathcal{L}_{base}=\mathcal{L}_{mask}+\mathcal{L}_{cls}$. $\mathcal{L}_{mask}$ stands for pixel-wise mask loss, which comprises of the BCE loss and the Dice loss. $\mathcal{L}_{cls}$ indicates the classification loss, where we use the cross-entropy loss. (2) the losses used to build inter-task associations, summarized as $\mathcal{L}_{inter}=\mathcal{L}_{v\to g}+\mathcal{L}_{v\to r}+\mathcal{L}_{con}+\mathcal{L}_{rank}$. The final loss function is $\mathcal{L}=\mathcal{L}_{base}+\lambda\mathcal{L}_{inter}$, where $\lambda$ is a balance weight, setting as $0.1$.

\section{Experiments}\label{sec:exp}

\textbf{Datasets.}
We evaluate the \ours~on three benchmarks: ScanNet20~\cite{dai2017scannet}, ScanNet200~\cite{rozenberszki2022lground}, and ScanRefer~\cite{chen2020scanrefer}. 
ScanNet20 provides RGB-D images and 3D point clouds of $1,613$ scenes, including $18$ instance categories and $2$ semantic categories. ScanNet200 uses the same source data as ScanNet20, while it is more challenging for 
up to $198$ instance categories and $2$ semantic categories. ScanRefer contains $51,583$ natural language expressions referring to $11,046$ objects selected from $800$ scenes. 

\textbf{Experimental setups.}
We train our method on the ScanNet20 training split, and the referring texts are collected from the ScanRefer. The $d_{in}$ and $d_{out}$ are set as $32$ and $256$, respectively. 
$m$ ranges $[50,100]$ percents of $M$ with an upper limit of $3,500$. We set $k$ as $10$ and $L$ as $6$.
For the data augmentations, the point clouds are randomly rotated around the z-axis, elastic distorted, and scaled; the referring texts are augmented using public GPT tools following~\cite{wu2023language,dai2023auggpt}. We adopt the AdamW optimizer with the polynomial schedule, setting an initial learning rate as $0.0001$ and the weight decay as $0.05$. All models are trained for $512$ epochs on a single NVIDIA RTX 4090 GPU and evaluated per $16$ epochs on the validation set to find the best-performed model. To stimulate the performance, we propose a two-stage fine-tuning trick, which fine-tunes the best-performed model using the learning rate and weight decay $0.001$ times the initial values for $40$ epochs. The proposed framework achieves end-to-end generic, interactive, and referring segmentation tasks. We divide the OVS task into mask prediction and class prediction. Specifically, we employ the proposed \ours~to predict masks and then follow the Open3DIS~\cite{nguyen2023open3dis} to generate class predictions.

We use PQ, mIoU, and mAP metrics to evaluate performance on the generic tasks following~\cite{narita2019panopticfusion,qi2017pointnet++,yi2019gspn}. Then, we use AP metric and mIoU for the interactive and referring segmentation tasks, respectively, following~\cite{yue2023agile3d, qian2024x}. For the OVS task, we train our model on the ScanNet20 and evaluate it using AP metric on the ScanNet200 without specific fine-tuning, following~\cite{takmaz2024openmask3d}. The Overall metric represents the average performance across six tasks intended to reflect the model's unified capability.

\setlength\tabcolsep{5.pt}
\begin{table*}
\small
\caption{Comparisons on ScanNet20~\cite{dai2017scannet}, ScanRefer~\cite{chen2020scanrefer}, and ScanNet200~\cite{rozenberszki2022lground}. The best results are highlighted in \textbf{bold}, and the second-best results are \underline{underscored}.
``$*$'' indicates the use of the two-stage fine-tuning trick. ``-/-'' denotes training on filtered or complete ScanRefer datasets.}
\begin{tabular*}{\linewidth}{@{\extracolsep{\fill}} lccccccc }
\toprule
\multicolumn{2}{c}{Datasets} & \multicolumn{4}{c}{ScanNet20} & \multicolumn{1}{c}{ScanRefer} & \multicolumn{1}{c}{ScanNet200} \\
\midrule
\multicolumn{2}{c}{3D scene understanding tasks}  & Pan. &Sem. & Inst. & Inter. & Ref. & OV \\
\midrule
Method &Reference & PQ & mIoU & mAP & AP & mIoU & AP \\ 
\midrule
SceneGraphFusion~\cite{wu2021scenegraphfusion} & CVPR 21 & 31.5 & - & - & - & - & - \\
TUPPer-Map~\cite{yang2021tuppermap} & IROS 21 & 50.2 & - & - & - & - & -   \\
Panoptic Lifting~\cite{siddiqui2023panoptic} & CVPR 23 & 58.9 & - & - & - & - & -   \\
PanopticNDT~\cite{seichter2023panopticndt} & IROS 23 & 59.2 & - & - & - & - & -   \\
\midrule
PointNeXt-XL~\cite{qian2022pointnext} & NeurIPS 22 & - & 71.5 & - & - & - & -   \\
PointMetaBase-XXL~\cite{lin2023pointmetabase} & CVPR 23 & - & 72.8 & - & - & - & -   \\
MM-3DScene~\cite{xu2023mm} & CVPR 23 & - & 72.8 & - & - & - & -   \\
PointTransformerV2~\cite{wu2022pointtransformerv2} & NeurIPS 22 & - & 75.4 & - & - & - & -   \\
ADS~\cite{Hong_2023_ICCV} & ICCV 23 & - & 75.6 & - & - & - & -   \\
OctFormer~\cite{wang2023octformer} & SIGGRAPH 23 & - & 75.7 & - & - & - & -   \\
\midrule
SoftGroup~\cite{vu2022softgroup} & CVPR 22 & - & - & 45.8 & - & - & -   \\

PBNet~\cite{zhao2023pbnet} & ICCV 23 & - & - & 54.3 & - & - & -   \\
ISBNet~\cite{ngo2023isbnet} & CVPR 23 & - & - & 54.5 & - & - & -   \\
SPFormer~\cite{sun2023superpoint} & AAAI 23 & - & - & 56.3 & - & - & -   \\
Mask3D~\cite{schult2023mask3d} & ICRA 23 & - & - & 55.2 & - & - & -   \\
MAFT~\cite{lai2023mask} & ICCV 23 & - & - & 58.4 & - & - & -   \\
QueryFormer~\cite{lu2023query} & ICCV 23 & - & - & 56.5 & - & - & -   \\
OneFormer3D~\cite{kolodiazhnyi2023oneformer3d} & CVPR 24 & \underline{71.2} & \underline{76.6} & \textbf{59.3} & - & - & -   \\
\midrule
InterObject3D~\cite{kontogianni2023interactive} & ICRA 23 & - & - & - & 20.9 & - & -   \\
AGILE3D~\cite{yue2023agile3d} & ICLR 24 & - & - & - & 53.5 & - & -   \\
\midrule
TGNN~\cite{huang2021text} & AAAI 21 & - & - & - & - & 24.9/27.8 & -   \\
X-RefSeg3D~\cite{qian2024x} & AAAI 24 & - & - & - & - & 25.5/29.9 & -   \\
\midrule
OpenScene~\cite{peng2023openscene} with~\cite{schult2023mask3d} & CVPR 23 & - & - & - & - & - & 8.5 \\
OpenMask3D~\cite{takmaz2024openmask3d} & NeurIPS 23 & - & - & - & - & - & 12.6 \\
SOLE~\cite{lee2024segment} & CVPR 24 & - & - & - & - & - & 18.7 \\
Open3DIS~\cite{nguyen2023open3dis} & CVPR 24 & - & - & - & - & - & 19.0 \\
\midrule
\ours~(\textbf{ours}) & - &\textbf{71.3}  & 76.3 & \underline{59.1} & \underline{54.1} & \underline{29.5}/- & \underline{19.6} \\
\ours$^{*}$~(\textbf{ours}) & - & \textbf{71.3} & \textbf{76.9} & \textbf{59.3} & \textbf{54.5} & \textbf{29.6}/- & \textbf{19.7} \\
\bottomrule
\end{tabular*}
\label{tab:comparison}
\end{table*}

\subsection{Comparison to the SOTA Methods}

The proposed method achieves six 3D scene understanding tasks in a single model. We demonstrate the effectiveness of our method by comparing it with SOTA approaches specialized for specific tasks.
As shown in Tab.~\ref{tab:comparison}, the proposed method outperforms the specialized SOTA methods PanopticNDT~\cite{seichter2023panopticndt}, OctFormer~\cite{wang2023octformer}, MAFT~\cite{lai2023mask}, AGILE3D~\cite{yue2023agile3d}, X-RefSeg3D~\cite{qian2024x}, and Open3DIS~\cite{nguyen2023open3dis} on the panoptic segmentation (PS), semantic segmentation (SS), instance segmentation (IS), interactive segmentation, referring segmentation, and OVS tasks by $12.1$ PQ, $1.2$ mIoU, $0.9$ mAP, $1.0$ AP, $4.1$ mIoU, $0.7$ AP, respectively. Even when compared with the competitive 3D unified method, \textit{i.e.}, OneFormer3D~\cite{kolodiazhnyi2023oneformer3d}, the proposed \ours~achieves $0.1$ PQ improvement on the PS task, and $0.3$ mIoU improvement on the SS task. More importantly, the OneFormer3D focuses on three generic segmentation tasks. It fails to understand user prompts and achieve OVS, which limits its application prospects. In contrast, \ours~unifies six tasks and presents desirable performance, demonstrating \ours~a powerful architecture.

\begin{wraptable}{b}{0.5\textwidth}
\centering
        \small
        \setlength{\tabcolsep}{1.2mm}
        \caption{Ablation on task unification.}
        \begin{tabular}{ cccccc }
            \toprule
            \multicolumn{1}{c}{ScanNet200} & \multicolumn{1}{c}{ScanRefer} & \multicolumn{4}{c}{ScanNet20} \\
            \midrule
            OV & Ref. & Inter. & Pan. &Sem. & Inst. \\
            \midrule
            AP & mIoU & AP & PQ & mIoU & mAP \\ 
            \midrule
            \ding{55} & \ding{55} & \ding{55} & \textbf{71.0} & 76.2 & \textbf{59.0} \\
            \ding{55} & \ding{55} & 56.8 & \textbf{71.0} & \textbf{76.4} & \underline{58.7} \\
            \ding{55} & 29.1 & 56.0 & 70.3 & \underline{76.3} & 58.4 \\
            19.7 & 29.1 & 54.5 & \underline{70.4} & 76.2 & 58.0 \\
            \bottomrule
        \end{tabular}
        \label{tab:ablation_task}
\end{wraptable}

The proposed method achieves six tasks in one training, which is elegant while facing an issue for fair comparison. Specifically, partial labels in the referring segmentation benchmark ({$10,115$ objects, $27.6\%$ of the complete ScanRefer training set}) annotate novel classes of the OVS task. Obviously, these labels should not be used for training to avoid label leakage. Thus, we filter out these labels and only employ the filtered ScanRefer training set to train our model. As shown in Tab.~\ref{tab:comparison}, our model uses $72.4\%$ training data to achieve closing performance with X-RefSeg3D~\cite{qian2024x} ($29.6$ vs. $29.9$), the current specialized SOTA on the 3D referring segmentation task. Moreover, while reproducing the X-RefSeg3D using official code on our filtered training data, the performance drops to $4.1$ mIoU lower than UniSeg3D,
demonstrating our model's effectiveness.

\subsection{Analysis and Ablation}\label{sec:ablation}
We conduct ablation studies and analyze the key insights of our designs. All models are evaluated on unified tasks to show the effectiveness of the proposed components on a broad scope.

\textbf{The challenge of multi-task unification.}
We first discuss the challenge of unifying multi-tasks in a single model. Specifically, we simply add interactive segmentation, referring segmentation, and OVS into our framework to build a unification baseline, as shown in Tab.~\ref{tab:ablation_task}. We observe a continuous performance decline on the PS, IS, and interactive segmentation tasks, indicating a significant challenge in balancing different tasks. Even so, we believe that unifying multiple tasks within a single model is worthy of exploring, as it can reduce computation consumption and benefit real-world applications. Thus, this paper proposes to eliminate performance decline by delivering inter-task associations, and the following experiments demonstrate that this could be a valuable step.

%

\setlength\tabcolsep{2.pt}
\begin{table*}[t]
\small
\caption{Ablation on components. ``Distillation'', ``Rank-Contrastive'', and ``Trick'' denote the knowledge distillation, ranking-based contrastive learning, and two-stage fine-tuning trick, respectively.}
\begin{tabular*}{\linewidth}{@{\extracolsep{\fill}} cccccccccc }
\toprule
\multicolumn{3}{c}{Datasets} & \multicolumn{4}{c}{ScanNet20} & \multicolumn{1}{c}{ScanRefer} & \multicolumn{1}{c}{ScanNet200} & \multirow{4}{*}{Overall} \\
\cmidrule{1-9}
\multicolumn{3}{c}{Components} & Pan. &Sem. & Inst. & Inter. & Ref. & OV & \\
\cmidrule{1-9}
Distillation & Rank-Contrastive & 
Trick & PQ & mIoU & mAP & AP & mIoU & AP & \\ 
\midrule
 - & - & - & 70.4 & 76.2 & 58.0 & \underline{54.5} & 29.1 & \underline{19.7} & 51.3 \\
\ding {52} & - & -  & \underline{70.9} & 76.2 & 58.6  & \textbf{55.3} & 29.2 & 19.6 & 51.6 \\
 - & \ding {52} & - & 70.8 & \underline{76.4} & 58.4 & 54.1 & \textbf{29.6} & \textbf{19.9} & 51.5 \\
\ding {52} & \ding {52} & - &\textbf{71.3}  & 76.3 & \underline{59.1} & 54.1 & \underline{29.5} & 19.6 & \underline{51.7} \\
\ding {52} & \ding {52} & \ding {52}  & \textbf{71.3} & \textbf{76.9} & \textbf{59.3} & \underline{54.5} & \textbf{29.6} & \underline{19.7} & \textbf{51.9} \\
\bottomrule
\end{tabular*}
\label{tab:ablation_component}
\end{table*}

\begin{table*}
    \small
    \centering
    \caption{Ablation on different designs of the proposed components. ``${v\to g}$'' and ``${v\to r}$'' denote the knowledge distillation from the interactive segmentation to the generic segmentation and the referring segmentation, respectively. ``Contrastive'' and ``Rank'' denote the contrastive learning and the ranking rule, respectively.}
    \label{tab:ablation_detail}
    \begin{subtable}[t]{1.0\linewidth}
        \centering
        \setlength{\tabcolsep}{3.25mm}
        \caption{Ablation on designs for knowledge distillation.}
        \begin{tabular}{cccccccccc}
            \toprule
            \multicolumn{2}{c}{Datasets} & \multicolumn{4}{c}{ScanNet20} & \multicolumn{1}{c}{ScanRefer} & \multicolumn{1}{c}{ScanNet200} & \multirow{4}{*}{Overall} \\
            \cmidrule{1-8}
            \multicolumn{2}{c}{Components} & Pan. &Sem. & Inst. & Inter. & Ref. & OV \\
            \cmidrule{1-8}
            ${v\to g}$ & ${v\to r}$ & PQ & mIoU & mAP & AP & mIoU & AP & \\ 
            \midrule
            - & - & 70.8 & \textbf{76.4} & 58.4 & \textbf{54.1} & \underline{29.6} & \underline{19.9} & 51.5\\
            \ding {52} & - & \underline{71.2} & \underline{76.3} & \underline{59.0} & \underline{54.0} & 29.5 & 19.8 & \underline{51.6}\\
            - & \ding {52} & 70.7 & 76.2 & 58.6 & \textbf{54.1} & \textbf{29.7} & \textbf{20.0} & \underline{51.6}\\
            \ding {52} & \ding {52} &\textbf{71.3}  & \underline{76.3} & \textbf{59.1} & \textbf{54.1} & 29.5 & 19.6 & \textbf{51.7}\\
            \bottomrule
        \end{tabular}
    \end{subtable}
    \begin{subtable}[t]{1.0\linewidth}
        \centering
        \setlength{\tabcolsep}{3.02mm}
        \caption{Ablation on designs for ranking-based contrastive learning.}
        \begin{tabular}{ccccccccc}
            \toprule
            \multicolumn{2}{c}{Datasets} & \multicolumn{4}{c}{ScanNet20} & \multicolumn{1}{c}{ScanRefer} & \multicolumn{1}{c}{ScanNet200} & \multirow{4}{*}{Overall} \\
            \cmidrule{1-8}
            \multicolumn{2}{c}{Components} & Pan. &Sem. & Inst. & Inter. & Ref. & OV \\
            \cmidrule{1-8}
            Contrastive & Rank & PQ & mIoU & mAP & AP & mIoU & AP & \\ 
            \midrule
            - & - & 70.9 & \underline{76.2} & 58.6  & \textbf{55.3} & 29.2 & 19.6 & \underline{51.6} \\
            \ding {52} & - & \underline{71.0}  & \textbf{76.3}  & \underline{59.0} & 54.5  & \underline{29.4} & \underline{19.7} & \textbf{51.7}\\
            - & \ding {52} & \underline{71.0} & \underline{76.2} & 58.7 & \underline{54.6} & \textbf{29.5} & \textbf{19.8} & \underline{51.6}\\
            \ding {52} & \ding {52} &\textbf{71.3}  & \textbf{76.3} & \textbf{59.1} & 54.1 & \textbf{29.5} & 19.6 & \textbf{51.7} \\
            \bottomrule
            
        \end{tabular}
    \end{subtable}

\end{table*}
\textbf{Design of inter-task associations.}
Our approach uses knowledge distillation and contrastive learning to connect supported tasks. As shown in Tab.~\ref{tab:ablation_component}, when applying the distillation, \textit{i.e.} row $2$, the performance of IS and interactive segmentation increase to $58.6$ mAP and $55.3$ AP, respectively. We believe the improvement on the IS task is because of the reliable knowledge distilled from the interactive segmentation, and the improvement on the interactive segmentation task is attributed to the intrinsic connections between the two tasks. Then, we ablate the ranking-based contrastive learning, \textit{i.e.} row $3$. We observe improvements on five tasks, including the generic segmentation and the referring segmentation, while a bit of performance drop on the interactive segmentation. This phenomenon suggests that contrastive learning is effective in most tasks, but there is a huge struggle to align the point and text modalities, which weakens the interactive segmentation performance.
Overall metric measures multi-task unification performance. We choose models and checkpoints with higher Overalls in our experiments. In practical applications, checkpoints can be chosen based on preferred tasks while maintaining good performance across other tasks.
Applying knowledge distillation and ranking-based contrastive learning obtains comparable performance on most tasks, performing higher Overall than rows $2$ and $3$, indicating the complementarity of the two components. We further employ two-stage fine-tuning trick, bringing consistent improvements across various tasks.

\begin{table*}
\small
 \centering
        \setlength{\tabcolsep}{3.38mm}
        \renewcommand\arraystretch{0.99}
        \caption{Ablation on hyper-parameter $\lambda$.}
        \begin{tabular}{cccccccc}
            \toprule
            \multicolumn{1}{c}{Datasets} & \multicolumn{4}{c}{ScanNet20} & \multicolumn{1}{c}{ScanRefer} & \multicolumn{1}{c}{ScanNet200} & \multirow{4}{*}{Overall} \\
            \cmidrule{1-7}
            Hyper-parameter & Pan. &Sem. & Inst. & Inter. & Ref. & OV & \\
            \cmidrule{1-7}
            $\lambda$ & PQ & mIoU & mAP & AP & mIoU & AP & \\ 
            \midrule
            0.05 & 70.7 & 76.2 & \underline{58.9} & \textbf{54.4} & 29.5 & \textbf{19.6} & \underline{51.6} \\
            0.1 &\textbf{71.3}  & \underline{76.3} & \textbf{59.1} & \underline{54.1} & 29.5 & \textbf{19.6} & \textbf{51.7} \\
            0.2 & \underline{70.8} & \textbf{76.6} & 58.6 & 52.3 & \textbf{29.8} & \underline{19.5} & 51.3\\
            0.3 & 70.6 & 75.7 & 58.4 & 51.6 & \underline{29.6} & 19.3 &50.9\\
            \bottomrule
        \end{tabular}
        \label{tab:hype_param}
\end{table*}

Detailed ablation on the components is shown in Tab.~\ref{tab:ablation_detail}. It is observed that knowledge distillation to various tasks brings respective improvements. As for contrastive learning, comparing row $2$ and row $4$ in Tab.~\ref{tab:ablation_detail}(b), the ranking rule suppresses the confusing point-text pairs, boosting contrastive learning to be more effective. $\lambda$ controls the strength of the explicit inter-task associations.
We empirically find that setting $\lambda$ to $0.1$ obtains the best performance, as shown in Tab.~\ref{tab:hype_param}.

\begin{wraptable}{r}{0.38\textwidth}
    \small
    \centering
    \captionsetup{type=table}
    \caption{Ablation on vision prompts.}
    \begin{tabular}{ cccccc }
        \toprule
        Strategy & mIoU & AP & AP$_{50}$ & AP$_{25}$ \\
        \midrule
        From~\cite{yue2023agile3d} & 78.8  & 54.5  & 79.4  & 93.2  \\
        Instance center & \textbf{79.6}  & \textbf{56.6}  & \textbf{82.1}  & \textbf{94.9}  \\
        $r_d=0.1$ & \underline{79.1}  & \underline{55.9}  & \underline{81.1}  & \underline{94.4}  \\
        $r_d=0.2$ & 78.7  & 55.1  & 80.0  & 93.4  \\
        $r_d=0.3$ & 78.0  & 53.8  & 78.5  & 92.4  \\
        $r_d=0.4$ & 77.5  & 53.0  & 77.4  & 91.7  \\
        $r_d=0.5$ & 76.6  & 52.1  & 76.2  & 90.6  \\
        $r_d=0.6$ & 75.9  & 51.2  & 74.6  & 90.0  \\
        $r_d=0.7$ & 74.9  & 50.1  & 72.9  & 88.1  \\
        $r_d=0.8$ & 73.4  & 48.2  & 71.1  & 86.5  \\
        $r_d=0.9$ & 71.0  & 45.3  & 66.6  & 82.1  \\
        $r_d=1.0$ & 62.7  & 36.4  & 54.8  & 70.2  \\
        Random & 76.0  & 51.3  & 75.2  & 89.6  \\
        \bottomrule
    \end{tabular}
    \label{tab:ablation_vision_prompt}
\end{wraptable}

\textbf{Influence of vision prompts.}
We empirically find that the vision prompts affect the interactive segmentation performance. To ensure a fair comparison, we adopt the same vision prompts generation strategy designed in AGILE3D~\cite{yue2023agile3d} to evaluate our interactive segmentation performance.

Furthermore, to analyze the influence of vision prompts,
we ablate the 3D spatial distances between the vision prompts and the instance centers. Specifically, assuming an instance containing $n$ points, we denote the mean coordinate of these points as the \textit{instance center} and order the $n$ points based on their distances to the instance center.
Then, we evaluate the interactive segmentation performance while using the $\left \lfloor r_d \times n \right \rfloor$-th nearest point as the vision prompt, as shown in Tab.~\ref{tab:ablation_vision_prompt}. When the vision prompt is located at the instance center, the interactive segmentation achieves the upper bound performance of $56.6$ AP. There is a significant performance gap (up to $20.2$ AP) between the edge and center points.
It illustrates considerable room for improvement. 
We observe an unusual decline in AP
while increasing $r_d$ from $0.9$ to $1.0$. We think this is because of the ambiguity in distinguishing the edge points from adjacent instances.
As we all know, this is the first work ablating the influence of vision prompts. We will explore it in depth in future work.

\section{Conclusion and Discussion}\label{sec:con}

We propose a unified framework named \ours, which provides a flexible and efficient solution for 3D scene understanding, supporting six tasks within a single model. Previous task-specific approaches fail to leverage cross-task information, limiting their understanding of 3D scenes to a task-specific perspective. In contrast, we take advantage of the multi-task design and enhance performance through building inter-task associations. Specifically, we employ knowledge distillation and ranking-based contrastive learning to facilitate cross-task knowledge sharing. Experiments demonstrate the proposed framework is a powerful method, achieving SOTA performance across six unified tasks.

\mysection{Limitation}\label{sec:limit} 
\ours~aims to achieve unified 3D scene understanding. However, it works on indoor tasks and lacks explorations in outdoor scenes. Additionally, we observe that \ours~performs worse interactive segmentation performance when the vision prompt is located away from the instance centers, limiting the reliability of the \ours~and should be explored in future work.

{\small
\bibliographystyle{plain}
\bibliography{main}
}

\newpage
\section*{Appendix}\appendix

\newcommand{\applabel}{Appendix\xspace}
\renewcommand{\thesection}{\Alph{section}}
\renewcommand{\thetable}{\Roman{table}}
\renewcommand{\thefigure}{\Roman{figure}}
\setcounter{section}{0}
\setcounter{table}{0}
\setcounter{figure}{0}

In this appendix, we provide additional content to complement the main manuscript:
\begin{itemize}[leftmargin=1em,topsep=0pt]
\item \applabel~\ref{sec:suppl_comparison}: Comparisons employing more metrics on specific tasks.
\item \applabel~\ref{sec:suppl_inference_time}: Inference time analysis of the proposed \ours.
\item \applabel~\ref{sec:suppl_visualization}: Qualitative visualizations illustrating model effectiveness.
\end{itemize}

\section{Comparisons employing more metrics on specific tasks.}
\label{sec:suppl_comparison}
The experiments presented in the main manuscript primarily use overarching metrics to measure performance on each task. This section provides more comprehensive comparisons of our method on each task using detailed metrics. We train the model on ScanNet20 and assess its open-vocabulary segmentation performance on ScanNet200. Following \cite{nguyen2023open3dis}, $51$ classes in ScanNet200 that are semantically similar to annotated classes in ScanNet20 are grouped as \texttt{Base} classes, while the remaining classes are divided as \texttt{Novel} classes. The model is then directly tested on Replica~\cite{straub2019replica} to evaluate its zero-shot segmentation performance.

\begin{figure}[ht]
        \small
    \centering
        \begin{minipage}{0.46\textwidth}
         \centering
        \captionsetup{type=table}
        \setlength{\tabcolsep}{0.5mm}
        \caption{Comparison with existing instance segmentation methods on ScanNet20. UniSeg3D achieves highly competitive performance.}
        \begin{tabular}{lccccc}
            \toprule
            Method & Reference & mAP\textsubscript{25} & mAP\textsubscript{50} & mAP \\
            \midrule
            3D-SIS\cite{hou20193dsis} & CVPR 19 & 35.7 & 18.7 & -\\
            GSPN\cite{yi2019gspn} & CVPR 19 & 53.4 & 37.8 & 19.3 \\ 
            PointGroup\cite{jiang2020pointgroup} & CVPR 20 & 71.3 & 56.7 & 34.8\\
            OccuSeg\cite{han2020occuseg} & CVPR 20 & 71.9 & 60.7 & 44.2 \\
            DyCo3D\cite{he2021dyco3d} & CVPR 21 & 72.9 & 57.6 & 35.4 \\
            SSTNet\cite{liang2021sstnet} & ICCV 21 & 74.0 & 64.3 & 49.4 \\
            HAIS\cite{chen2021hais} & ICCV 21 & 75.6 & 64.4 & 43.5 \\
            DKNet\cite{wu2022dknet} & ICCV 22 & 76.9 & 66.7 & 50.8 \\
            SoftGroup\cite{vu2022softgroup} & CVPR 22 & 78.9 & 67.6 & 45.8 \\
            PBNet\cite{zhao2023pbnet} & ICCV 23 & 78.9 & 70.5 & 54.3 \\
            ISBNet\cite{ngo2023isbnet} & CVPR 23 & 82.5 & 73.1 & 54.5 \\
            SPFormer\cite{sun2023superpoint} & AAAI 23 & 82.9 & 73.9 & 56.3 \\
            Mask3D\cite{schult2023mask3d} & ICRA 23 & 83.5 & 73.7 & 55.2 \\
            MAFT\cite{lai2023mask} & ICCV 23 & - & 75.9 & \underline{58.4} \\
            QueryFormer\cite{lu2023query} & ICCV 23 & 83.3 & 74.2 & 56.5 \\
            OneFormer3D\cite{kolodiazhnyi2023oneformer3d} & CVPR 24 & \textbf{86.4} & \textbf{78.1} & \textbf{59.3} \\
            \ours~(\textbf{ours}) & - & \underline{86.1} & \underline{77.0} & \textbf{59.3}\\
            \bottomrule
        \end{tabular}
        \label{tab:scannet} 
    \end{minipage}
    \hfill
    \begin{minipage}{0.51\textwidth}
        \begin{minipage}{1\linewidth}
             \centering
            \captionsetup{type=table}
            \setlength{\tabcolsep}{1.6mm}
            \caption{Comparison with previous 3D interactive segmentation methods on ScanNet20. UniSeg3D presents remarkable performance in terms of three metrics.}
            \begin{tabular}{lcccc}
                \toprule
                Method & Reference & AP & AP$_{\text{50}}$ & AP$_{\text{25}}$ \\ 
                \midrule
                InterObject3D~\cite{kontogianni2023interactive} & ICRA 23 & 20.9 & 38.0 & 67.2 \\
                AGILE3D~\cite{yue2023agile3d} & ICLR 24 & \underline{53.5} & \underline{75.6} & \underline{91.3} \\
                \ours~(\textbf{ours}) & - & \textbf{54.5} & \textbf{79.4} & \textbf{93.2} \\
                \bottomrule
            \end{tabular}
            \label{tab:interactive_segmentation} 
        \end{minipage}
        \begin{minipage}{1.\textwidth}
            \noindent
            \centering
            \small
            \captionsetup{type=table}
                \setlength{\tabcolsep}{0.5mm}
                \caption{Comparison with existing 3D referring segmentation methods on ScanRefer. UniSeg3D demonstrates notable performance in terms of mIoU and acc@0.25.}
                \begin{tabular}{lcccc}
                    \toprule
                    Method & Reference & mIoU & acc@0.5 & acc@0.25 \\ 
                    \midrule
                    TGNN~\cite{huang2021text} & AAAI 21 & 24.9 & \underline{28.2} & 33.2 \\
                    X-RefSeg3D~\cite{qian2024x} & AAAI 24 & \underline{25.5} & \textbf{28.6} & \underline{34.0} \\
                    \ours~(\textbf{ours}) & - & \textbf{29.6} & 28.0 & \textbf{41.5} \\
                    \bottomrule
                \end{tabular}
                \label{tab:scanrefer}  
    \end{minipage} 
    \end{minipage}
\end{figure}

\begin{table*}[ht]
\centering
    \small
    \captionsetup{type=table}
        \setlength{\tabcolsep}{3.05mm}
        \caption{Comparison with previous open-vocabulary segmentation methods on ScanNe200 and Replica. Our method outperforms existing approaches in terms of AP.}
        \begin{tabular}{lccccccc}
            \toprule
            \multirow{2.5}{*}{Method} & \multirow{2.5}{*}{Reference} & \multicolumn{3}{c}{ScanNet200} & \multicolumn{3}{c}{Replica} \\
            \cmidrule{3-8}
             &  & AP & AP$_{\texttt{Base}}$ & AP$_{\texttt{Novel}}$ & AP & AP$_{50}$ & AP$_{25}$ \\ 
            \midrule
            OpenScene~\cite{peng2023openscene} with~\cite{schult2023mask3d} & CVPR 23 & 8.5 & 11.1 & 7.6 & 10.9 & 15.6 & 17.3 \\
            OpenMask3D\cite{takmaz2024openmask3d} & NeurIPS 23 & 12.6 & 14.3 & 11.9 & 13.1 & 18.4 & 24.2 \\
            SOLE\cite{lee2024segment} & CVPR 24 & 18.7 & 17.4 & \textbf{19.1} & - & - & -\\
            Open3DIS\cite{nguyen2023open3dis} & CVPR 24 & \underline{19.0} & \textbf{25.8} & 16.5 & \underline{18.5} & \textbf{24.5} & \underline{28.2} \\
            \ours~(\textbf{ours}) & - & \textbf{19.7} & \underline{24.4} & \underline{18.0} & \textbf{19.1} & \underline{24.1} & \textbf{29.2} \\
            \bottomrule
        \end{tabular}
        \label{tab:ov}
\end{table*}

\clearpage

\section{Inference time analysis of the proposed \ours.}
\label{sec:suppl_inference_time}

This work proposes a unified framework, achieving six tasks in one inference, which would be more efficient than running six task-specific approaches individually. We present the inference time of the proposed method for efficiency analysis. 
Tab.~\ref{tab:profiler} illustrates that our method achieves effective integration across six tasks while maintaining highly competitive inference times compared to previous methods. 
\begin{table*}[ht]
\centering
\small
\setlength{\tabcolsep}{1.2mm}
\caption{Inference time and instance segmentation performance on the ScanNet20 validation split.}
\begin{tabular}{llcccc}
\toprule
\multirow{2}{*}{Method} & \multirow{2}{*}{Component} & \multirow{2}{*}{Device} & \multirow{2}{*}{\shortstack{Component\\time, ms}} & \multirow{2}{*}{\shortstack{Total\\time, ms}} & \multirow{2}{*}{mAP} \\
& & & & & \\
\midrule
\multirow{3}{*}{PointGroup~\cite{jiang2020pointgroup}} & Backbone & GPU & 48 & \multirow{3}{*}{372} & \multirow{3}{*}{34.8} \\
& Grouping & GPU+CPU & 218 & & \\
& ScoreNet & GPU & 106 & & \\
\midrule
\multirow{3}{*}{HAIS~\cite{chen2021hais}} & Backbone & GPU & 50 & \multirow{3}{*}{256} & \multirow{3}{*}{43.5} \\
& Hierarchical aggregation & GPU+CPU & 116 & & \\
& Intra-instance refinement & GPU & 90 & & \\
\midrule
\multirow{3}{*}{SoftGroup~\cite{vu2022softgroup}} & Backbone & GPU & 48 & \multirow{3}{*}{266} & \multirow{3}{*}{45.8} \\
& Soft grouping & GPU+CPU & 121 & & \\
& Top-down refinement & GPU & 97 & & \\
\midrule
\multirow{4}{*}{SSTNet~\cite{liang2021sstnet}} & Superpoint extraction & CPU & 168 & \multirow{4}{*}{400} & \multirow{4}{*}{49.4} \\
& Backbone & GPU & 26 & & \\
& Tree Network & GPU+CPU & 148 & & \\
& ScoreNet & GPU & 58 & & \\
\midrule
\multirow{3}{*}{\shortstack[l]{Mask3D~\cite{schult2023mask3d}\\w/o clustering}} & Backbone & GPU & 106 & \multirow{3}{*}{221} & \multirow{3}{*}{54.3} \\
& Mask module & GPU & 100 & & \\
& Query refinement & GPU & 15 & & \\
\midrule
\multirow{4}{*}{Mask3D~\cite{schult2023mask3d}} & Backbone & GPU & 106 & \multirow{4}{*}{19851} & \multirow{4}{*}{55.2} \\
& Mask module & GPU & 100 & & \\
& Query refinement & GPU & 15 & & \\
& DBSCAN clustering & CPU & 19630 & & \\  
\midrule
\multirow{4}{*}{SPFormer~\cite{sun2023superpoint}} & Superpoint extraction & CPU & 168 & \multirow{4}{*}{215} & \multirow{4}{*}{56.3} \\
& Backbone & GPU & 26 & & \\
& Superpoint pooling & GPU & 4 & & \\
& Query decoder & GPU & 17 & & \\  
\midrule
\multirow{4}{*}{OneFormer3D~\cite{kolodiazhnyi2023oneformer3d}} & Superpoint extraction & CPU & 168 & \multirow{4}{*}{221} & \multirow{4}{*}{59.3} \\
& Backbone & GPU & 26 & & \\
& Superpoint pooling & GPU & 4 & & \\
& Query decoder & GPU & 23 & & \\  
\midrule
\multirow{4}{*}{\ours~(\textbf{ours})} & Superpoint extraction & CPU & 168 & \multirow{4}{*}{230.03} & \multirow{4}{*}{59.3} \\
& Backbone & GPU & 33 & & \\
& Text encoder & GPU & 0.03 & & \\
& Mask decoder & GPU & 29 & & \\ 
\bottomrule
\end{tabular} 
\label{tab:profiler}
\end{table*}


\section{Qualitative visualizations illustrating model effectiveness.}
\label{sec:suppl_visualization}

\begin{figure}[ht] 
\centering
\includegraphics[width=0.98\textwidth]{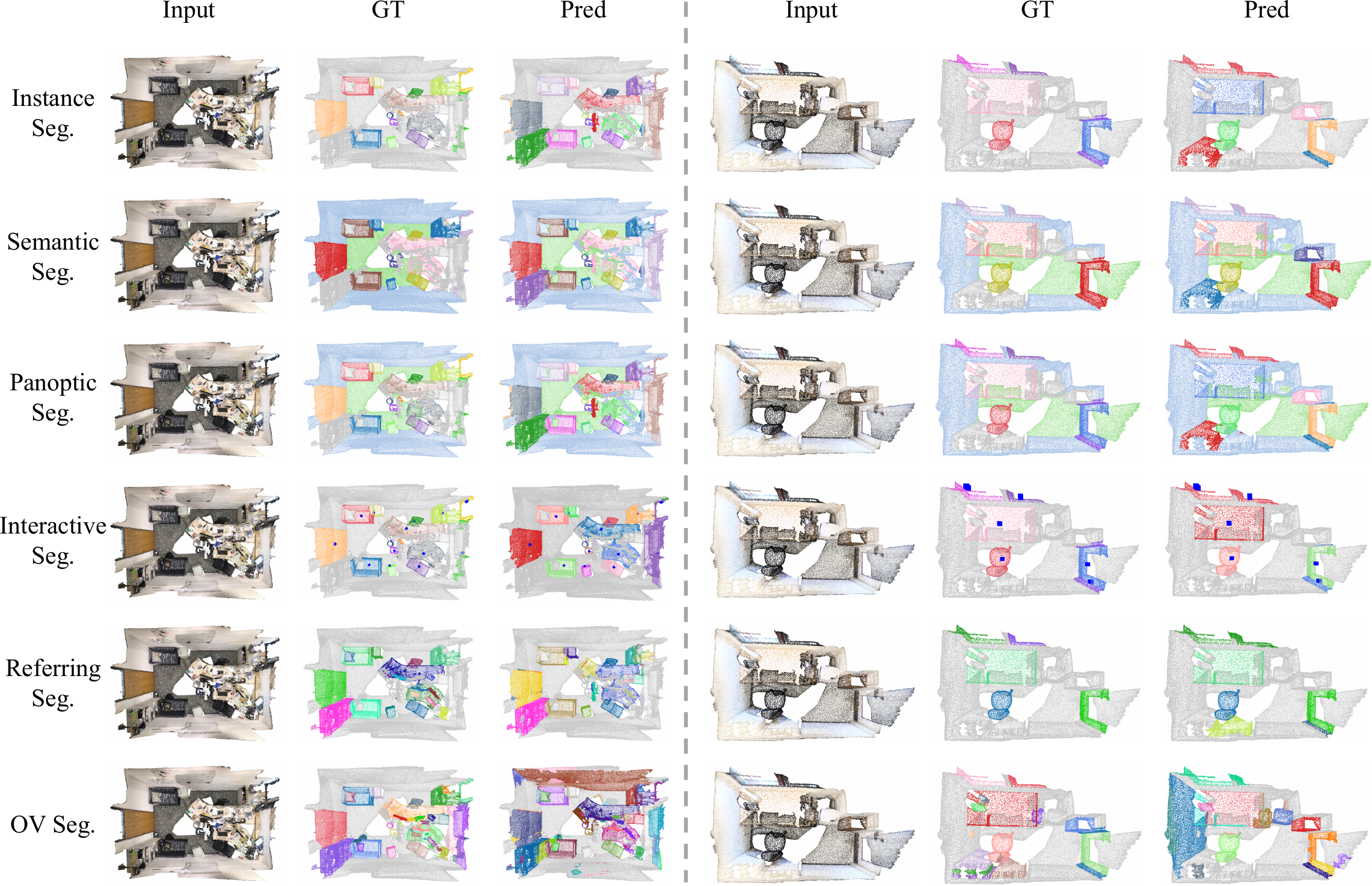}
\caption{Visualization of segmentation results obtained by UniSeg3D on ScanNet20 validation split.
}
\label{fig:visualization}
\end{figure}

\begin{figure}[ht] 
\centering
\includegraphics[width=0.98\textwidth]{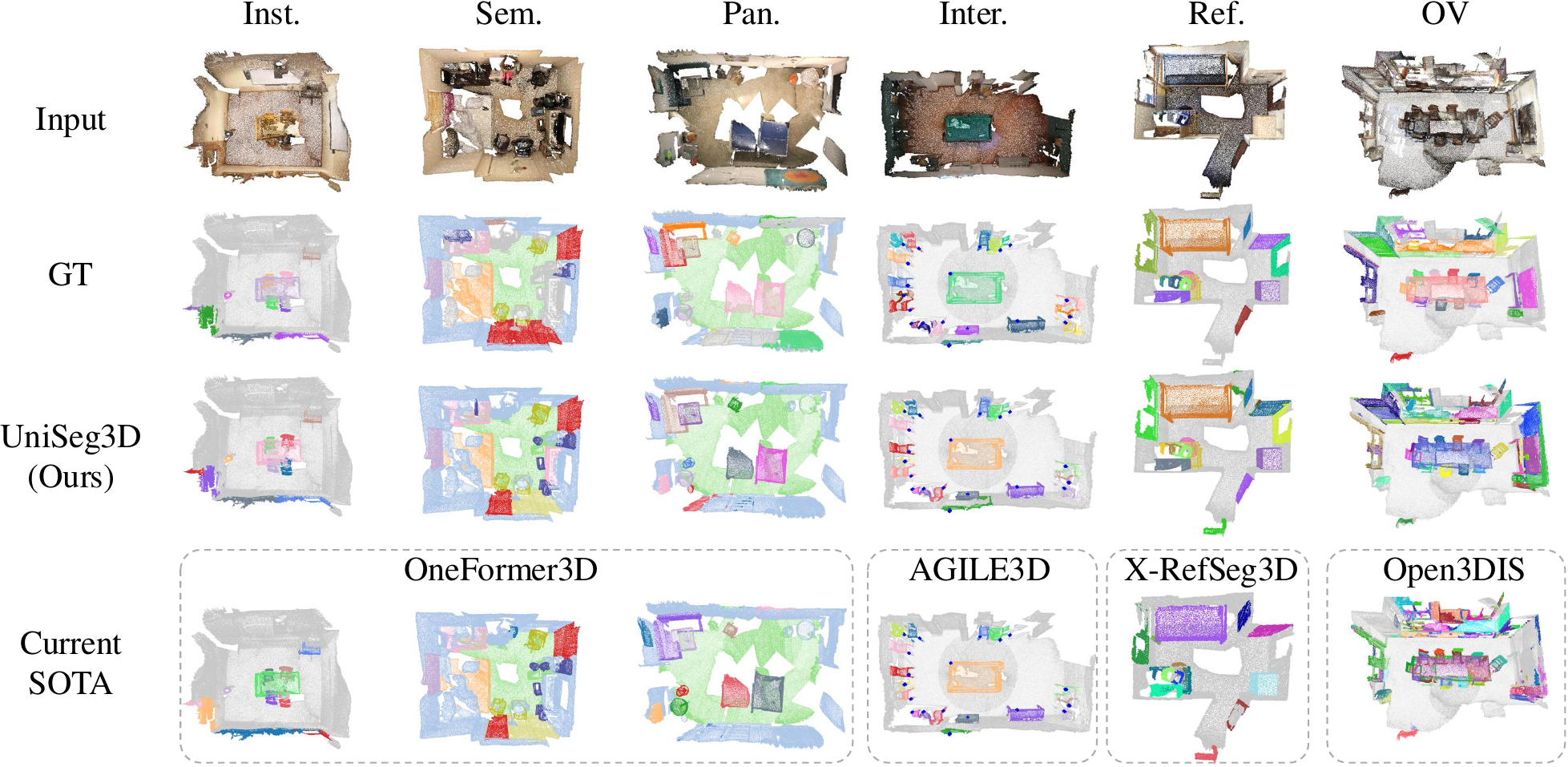}
\caption{Visualization of segmentation results obtained by UniSeg3D and current SOTA methods on ScanNet20 validation split.
}
\label{fig:visualization_comparison}
\end{figure}

\begin{figure}[t] 
\centering
\includegraphics[width=0.98\textwidth]{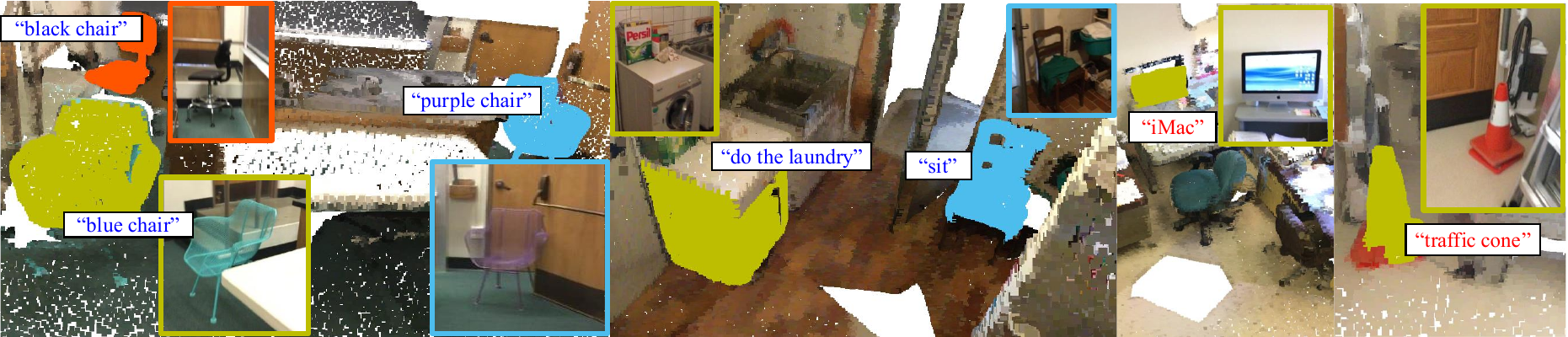}
\caption{Visualization of open capabilities. \textcolor{red}{Red prompts} involve categories not presented in the ScanNet200 annotations, while \textcolor{blue}{blue prompts} describe the attributes of various objects, such as affordances and color.
}
\label{fig:visualization_resoning}
\end{figure}

We provide qualitative results in this section. In Fig.~\ref{fig:visualization}, visualizations of multi-task segmentation results are presented, showcasing point clouds, ground truth, and predictions within each scene. In Fig.~\ref{fig:visualization_comparison}, we present visualizations of predictions from \ours~and current SOTA methods.
In Fig.~\ref{fig:visualization_resoning}, we test our model on open-set classes not included in training data to evaluate the model's open capability. Furthermore, we even replace the class names with attribute descriptions in the open vocabulary, and impressively, we observe the preliminary reasoning capabilities of our approach.

\end{document}